\documentclass[lettersize,journal]{IEEEtran}
\usepackage{amsmath,amsfonts}
\usepackage{algorithmic}
\usepackage{algorithm}
\usepackage{array}
\usepackage[caption=false,font=normalsize,labelfont=sf,textfont=sf]{subfig}
\usepackage{textcomp}
\usepackage{stfloats}
\usepackage{url}
\usepackage{verbatim}
\usepackage{graphicx}
\usepackage{cite}
\usepackage[numbers,sort&compress]{natbib}
\usepackage{graphicx}%
\usepackage{multirow}%
\usepackage{amsthm}%
\usepackage{mathrsfs}%
\usepackage{xcolor}%
\usepackage{textcomp}%
\usepackage{manyfoot}%
\usepackage{booktabs}%
\usepackage{listings}%
\usepackage{caption}
\usepackage{hyperref}

\begin{document}


\title{Continuum-Interaction-Driven Intelligence: Human-Aligned Neural Architecture via Crystallized Reasoning and Fluid Generation}

\author{\IEEEauthorblockN{
    Pengcheng Zhou\IEEEauthorrefmark{1}, 
    Zhiqiang Nie\IEEEauthorrefmark{1},
    Haochen Li\IEEEauthorrefmark{1}}
    \\ \IEEEauthorblockA{\IEEEauthorrefmark{1}Department of Computer Science and Technology, Tsinghua University}
}

\maketitle

\begin{abstract}
Current AI systems based on probabilistic neural networks, such as large language models (LLMs), have demonstrated remarkable generative capabilities yet face critical challenges including hallucination, unpredictability, and misalignment with human decision-making. These issues fundamentally stem from the over-reliance on randomized (probabilistic) neural networks—oversimplified models of biological neural networks—while neglecting the role of procedural reasoning (chain-of-thought) in trustworthy decision-making. Inspired by the human cognitive duality of fluid intelligence (flexible generation) and crystallized intelligence (structured knowledge) \cite{cattell1963theory}, this study proposes a dual-channel intelligent architecture that integrates probabilistic generation (LLMs) with white-box procedural reasoning (chain-of-thought) to construct interpretable, continuously learnable, and human-aligned AI systems.

Concretely, this work: (1) redefines chain-of-thought as a programmable crystallized intelligence carrier, enabling dynamic knowledge evolution and decision verification through multi-turn interaction frameworks; (2) introduces a task-driven modular network design that explicitly demarcates the functional boundaries between randomized generation and procedural control to address trustworthiness in vertical-domain applications; (3) demonstrates that multi-turn interaction is a necessary condition for intelligence emergence, with dialogue depth positively correlating with the system's human-alignment degree. This research not only establishes a new paradigm for trustworthy AI deployment but also provides theoretical foundations for next-generation human-AI collaborative systems.

{\itshape \textcolor{gray}{Notably, the framework positions multi-turn interaction as the core substrate of intelligence emergence—contrasting with the conventional "command-response" ant paradigm (i.e., mechanical task decomposition and execution). Sustained interaction achieves cognitive leaps through: (a) temporal accumulation of cognitive states forming a human-like working memory buffer; (b) dynamic knowledge representation calibration via error backpropagation across interaction turns. This paradigm proves that authentic domain intelligence must be built upon interaction continuity rather than discrete command concatenation.}}

\end{abstract}


\section{Introduction}
The cognitive marvel of the human brain stems from its unique neural computational architecture: randomly connected synaptic networks in the neocortex support probabilistic pattern recognition \cite{fiser2010statistically,lee2003hierarchical}, while closed-loop circuits in the basal ganglia enable procedural skill consolidation \cite{friston2010free,graybiel1998basal,miller2001integrative,graybiel2005basal,doya1999computations,kahneman2011thinking,booch2021thinking}. This dual-channel mechanism, however, suffers from striking decoupling in artificial intelligence—all current knowledge injection paradigms, whether labeled knowledge in supervised learning \cite{lecun2015deep}, data-implied knowledge in self-supervised learning \cite{devlin2019bert}, or interactive knowledge acquisition in reinforcement learning \cite{sutton1998reinforcement}, focus solely on simulating the neocortex's probabilistic learning function\cite{lake2017building,marcus2020next}. 

This reductionism creates a fundamental paradox: the more we refine neural networks' knowledge injection mechanisms (e.g., reward model design in RLHF \cite{ouyang2022training}), the more we expose systemic flaws in vertical applications. Essentially, these methods attempt to simulate human multi-channel knowledge processing through a single probabilistic channel (gradient descent-guided parameter space search \cite{lecun2015deep}), , akin to expecting a disintegrated neural assembly \cite{tononi2008neural} to sustain coherent conscious processing—a computational fallacy.

This architectural deficiency manifests as a triple alienation during knowledge injection: First, supervised learning compresses human knowledge into discrete labels, destroying the topological structure of knowledge ontology (comparable to preserving the hippocampus' pattern separation while losing pattern completion \cite{mcclelland1995there}). Second, self-supervised learning retains data-inherent relationships but confines knowledge extraction to statistical significance, failing to replicate human top-down active inference (e.g., prefrontal modulation of sensory cortices \cite{clark2013whatever}). Third, reinforcement learning achieves gradual knowledge updates through environmental interaction \cite{mnih2015human}, yet its Markov decision process fundamentally differs from human episodic memory consolidation—whereas the brain reorganizes discrete experiences into narrative memory via hippocampal indexing \cite{eichenbaum2017memory,hassabis2017neuroscience}, RL agents can only perform isolated policy optimization. These alienations ultimately cause even human-level models to produce hallucinated outputs violating professional common sense in structured reasoning tasks like medical diagnosis or legal argumentation \cite{marcus2020next}.

We reveal a counterintuitive insight: The challenge in vertical domains stems not from insufficient knowledge injection but from fundamental misrepresentation \cite{lake2017building}. During expert decision-making, human reasoning concurrently activates two neural representations: probabilistic hypotheses from neocortical distributed networks (analogous to LLM generation \cite{vaswani2017attention}), and procedural checklists maintained by basal ganglia-thalamic circuits. Existing systems implement only the former \cite{liu2025advances}, degrading professional knowledge into statistical correlations rather than causal frameworks \cite{mccoy2019right}.

Concretely, AI's core challenges originate from its singular reliance on probabilistic networks \cite{pearl2018book}: First, LLM hallucinations reflect neural computational dysregulation—without basal ganglia-like verification circuits \cite{graybiel2005basal}, uncontrolled probability sampling during forward propagation activates erroneous knowledge nodes, causing catastrophic errors (e.g., prescribing contraindicated drugs \cite{topol2019high}). Second, the black-box problem maps to disrupted knowledge topology locality \cite{olah2020zoom}: Human experts pinpoint neuroanatomical correlates of decisions (e.g., temporal cortex for legal code recall), whereas LLM knowledge disperses diffusely across billions of parameters, preventing deterministic knowledge localization. Most critically, cognitive misalignment—between AI-human \cite{bai2022training} and intra-AI systems \cite{chenreasoning,liu2023cognitive,mondal2024large,ye2022unreliability}—exposes reasoning as post hoc probability reconstruction rather than genuine logic. This compels us to reconsider: are the architectural limitations of neural networks rooted in their oversimplified theoretical foundations regarding biological neural mechanisms?

Since the McCulloch-Pitts model (1943) abstracted neurons as binary threshold logic units \cite{mcculloch1943logical}, traditional artificial neural networks have long adopted cascaded weighted sums and nonlinear activations as their basic computational paradigm, while neglecting: dendritic sublinear computation capabilities \cite{gidon2020dendritic}, spike-timing-dependent synaptic plasticity \cite{bi1998synaptic}, and diffuse neuromodulatory regulation \cite{rice2008dopamine}. Although neuroscience has revealed multi-scale computational principles, mainstream deep learning still largely inherits early simplified modeling traditions. The limitations of this theoretical framework may partially explain the challenges AI systems face in specific cognitive tasks. When large language models generate professionally nonsensical hallucinations, or reinforcement learning agents fail to achieve human-like skill transfer, are we witnessing systemic defects caused by missing biological computation principles? The more profound question is: Can artificial neural networks relying solely on statistical correlation learning ever achieve true brain-like intelligence?

These defects collectively constitute the "glass ceiling" for AI adoption in vertical domains: in the field of healthcare, one of the main reasons for the low clinical adoption rate of top-tier diagnostic models is that physicians are unable to verify their reasoning pathways \cite{kelly2019key}; in legal consulting, professional users reject AI-generated contract clauses due to their inability to demonstrate complete logical chains of element review like human lawyers \cite{surden2018artificial,zafar2024balancing}. More ironically, current multi-agent systems optimized through reinforcement learning actually amplify these flaws—lacking prefrontal cortex-like consensus mechanisms \cite{miller2001integrative}, inter-agent interactions often cause error cascade propagation \cite{wu2023models}. This dilemma reveals a harsh reality: advancing AI merely by scaling parameters and improving training algorithms without constructing correct neural architectures is like attempting to treat Alzheimer's disease by increasing neuron count—it ignores that cognitive function depends fundamentally on neural network organization rather than scale \cite{tononi1994measure}.

Our Neuromorphic Intelligence Architecture (NIA) addresses this crisis by reconstructing the knowledge injection pathway: while receiving traditional training data, the system simultaneously constructs: (1) a probabilistic knowledge base (simulating the neocortex), (2) procedural verification rules (simulating the basal ganglia), and (3) a dialogic memory reorganization mechanism (simulating the hippocampus). This triple representation ensures that professional knowledge maintains a human-compatible structured format upon injection. For example, when applied in the legal field, legal texts are not only transformed into word vectors but also automatically generate procedural knowledge such as flowchart for element review and templates for evidence chain construction. These structured representations will be dynamically invoked and revised in subsequent dialogues.

Meanwhile, the dialogue engine of NIA breaks through the ant paradigm through three mechanisms: (1) the dialogue state tracker (simulating the prefrontal cortex) maintains cognitive consistency across turns; (2) the feedback integrator (simulating the thalamic reticular nucleus) enables the propagation and correction of errors over the dialogue timeline; (3) the intention generator (simulating the default mode network) proactively initiates clarifying inquiries. This design enables the system to gradually approach the core of the problem through a "questioning-clarifying-verifying" dialogue spiral, just like human experts, rather than exhausting all decision possibilities in a single-round interaction like traditional task-oriented AI.

The revolutionary aspect of this architecture lies in its transformation of the traditional AI mode of "knowledge solidification after injection" into a living system of "knowledge development upon injection." Taking the medical scenario as an example, when the system receives a new version of the treatment guidelines, a traditional Large Language Model (LLM) requires retraining the entire model (which is equivalent to erasing old memories and then reconstructing them), while the NIA system can complete the knowledge update through multiple rounds of professional dialogue—this process precisely simulates the way human experts integrate new knowledge through academic discussions. More crucially, all knowledge updates leave auditable cognitive traces, completely resolving the application barriers of black-box models in compliance-sensitive fields.

The ultimate implication of this research is that, in order to achieve true intelligence of AI systems in specific fields, more attention should be paid to the development of domain-adaptive neural computational structures, rather than simply relying on the injection of "general knowledge." Just as the human brain combines specialized regions (like the hippocampus for memory) with general cortical processing \cite{graybiel2005basal}, AI systems designed for vertical domains may require a dual-channel architecture: one channel is used to absorb domain-specific knowledge, and the other channel focuses on internalizing the logic and methodology of that domain. Through this targeted design approach, the limitations of large-scale models in practical applications can be more effectively addressed, providing a biologically reasonable strategy to advance the development of AI technology.

Specifically, this work: (1) redefines the chain of thought as a programmable carrier of crystallized intelligence, enabling dynamic knowledge evolution and decision-making verification through a multi-round dialogue framework, which is jointly maintained by humans and machines; (2) proposes a task-driven modular network design, clearly defining the functional boundaries between randomized generation and process-oriented control to address the credibility issue of implementation in vertical domains; (3) demonstrates that multi-round interaction is a necessary condition for the emergence of intelligence, and there is a positive correlation between the depth of dialogue and the degree of human alignment of the system. This architecture significantly outperforms traditional single probability models in reducing the hallucination rate and enhancing decision predictability. This research not only provides a new paradigm for the trustworthy implementation of AI but also lays a theoretical foundation for the design of the next generation of human-machine collaborative systems.

\begin{figure*}[htbp]
    \centering
    \includegraphics[width=\textwidth]{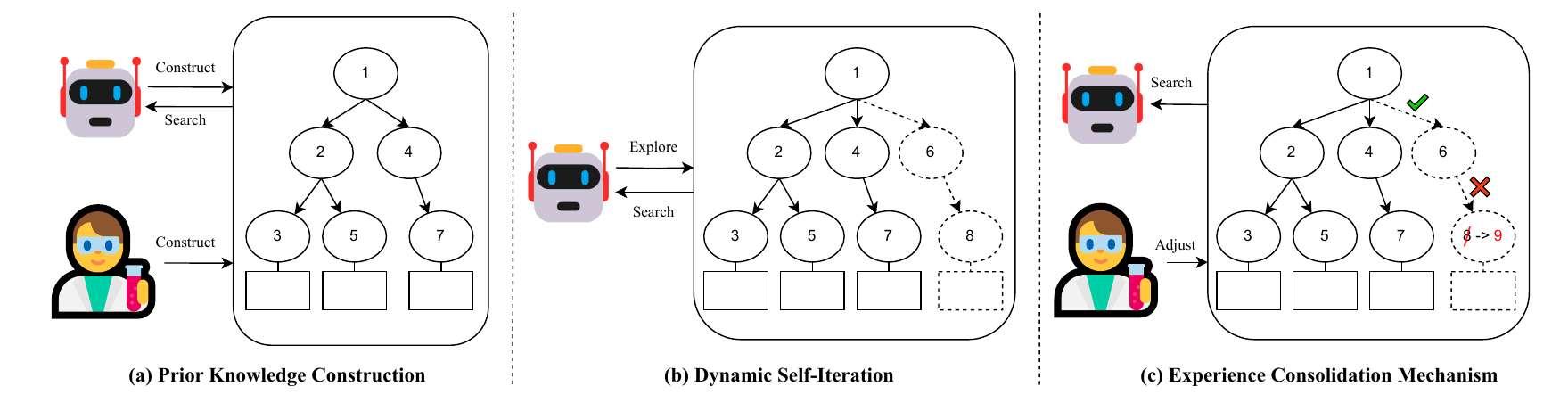}
    \caption{Crystallized Reasoning.}
    \label{fig:method}
\end{figure*}

\section{Methods}
This work proposes a dual-channel neural architecture that integrates proceduralized chain-of-thought (crystallized intelligence) with probabilistic neural networks (fluid intelligence), aiming to construct an interpretable, iterative, and human-aligned intelligent system. The methodology comprises three core modules, corresponding respectively to the knowledge initialization, learning process, and experience consolidation stages of human cognition.

\subsection{Neural Network (Chain-of-Thought) Initialization: Structured Prior Knowledge Construction}

The initialization of neural network based chain-of-thought aims to provide structured prior knowledge for subsequent learning, with the core challenge lying in balancing data - driven induction capability and logic - constrained controllability. We propose two possible construction paths: multi - agent consensus based case extraction and human - prescribed deterministic rule initialization. 

\subsubsection{Multi-Agent Consensus Based Case Extraction}
The essence of this method lies in the automated extraction of high-confidence reasoning chains from a large number of real-world cases. These reasoning chains are presented in a tree-like structure, clearly illustrating the logic and steps involved in multi-turn interactions. Specifically, we achieve this goal by deploying multiple agents, which collaborate rather than operate independently. Each agent is responsible for extracting candidate reasoning chains from the case set, explicitly defining the tasks to be performed at each step, and forming a comprehensive tree-like structure. After extraction, human experts verify, modify, and adjust the generated reasoning chains to ensure they meet high standards of reliability.

Specifically, during the case parsing and preliminary extraction phase, we deploy a dedicated agent $A$ to process the input case set $D = \{(x_i, y_i)\}_{i=1}^N$, where each multi-turn interaction case $(x_i, y_i)$ contains multi-turn input information $x_i$ and corresponding multi-turn output or results $y_i$. The task of agent $A$ is to extract candidate reasoning chains $C_i$ from each case. These reasoning chains are represented as tree structures, where each node corresponds to a reasoning step or decision point, and edges represent the logical relationships between steps. To ensure the accuracy of extraction, the agent system employs differentiated prompting strategies. For example, depending on the type of task or case characteristics, the agent dynamically adjusts its prompting templates to better capture the diverse information within the cases. Ultimately, the agent generates $N$ candidate reasoning chains $\{C_1, C_2, \dots, C_N\}$, with each chain corresponding to one case.

Since the reasoning chains extracted from different cases may exhibit variations in expression, directly merging them could lead to logical inconsistencies or redundancies. Therefore, we need to design a mechanism to semantically align and merge these reasoning chains. For each node in the reasoning chains, we represent its meaning using a semantic embedding vector $f(u)$. The semantic similarity between two nodes $u$ and $v$ can be calculated using cosine similarity:
\[
\text{sim}(u, v) = \cos(f(u), f(v)).
\]
By comparing all nodes in the reasoning chains pairwise, we identify semantically similar nodes and align them into a single logical unit. For instance, if multiple reasoning chains contain semantically similar reasoning steps, these steps can be merged into a universal node.

After completing node alignment, we aggregate the paths across different reasoning chains. For each pair of nodes $(u, v)$, if there are multiple logical paths between them, we select the optimal path based on confidence scores or other evaluation metrics. The confidence score is calculated using the following formula:
\[
\bar{P}(e) = \frac{1}{|K_e|} \sum_{k \in K_e} P_k(\phi^{-1}(e)),
\]
where $K_e$ represents the set of reasoning chains supporting path $e$, $\phi^{-1}(e)$ is the mapping of path $e$ in the corresponding reasoning chain, and $P_k(\cdot)$ is the confidence score of that path in reasoning chain $k$. Through this approach, the system can filter out the most reliable logical paths.

After path aggregation, the system performs a global consistency check on the entire reasoning chain to ensure there are no logical contradictions or cyclic dependencies. For example, certain paths may appear reasonable locally but could lead to conflicts from a global perspective. In such cases, the system prioritizes retaining high-confidence paths while discarding low-confidence or redundant ones.

After the above steps, the system generates a canvas-style logical chain $G$. However, since automated methods may introduce errors or omissions, human experts are required to verify and optimize the generated reasoning chain. Experts modify and adjust the chain based on domain knowledge and practical requirements, ensuring its logical rigor and alignment with real-world application scenarios. The optimized reasoning chain will serve as the standard for subsequent task execution.

Through this process, we can effectively merge multiple reasoning chains extracted from $n$ cases into a single high-quality, logically consistent unified reasoning chain. This method not only addresses the challenge of path merging in multi-case extraction but also leverages the expertise of human experts, laying a solid foundation for the efficient execution of subsequent tasks.

\subsubsection{Human-Prescribed Deterministic Rule Initialization}
In the process of experts directly constructing reasoning chains, a foundational rule set $R_{\text{atomic}} = \{ r_j \}_{j=1}^M$ is first defined. These rules include hard constraints and are applicable to various common scenarios. Each rule precisely describes the actions or decision logic to be taken under specific conditions.

Next, different atomic rules are combined into a canvas-style logical chain $G$ through structural grafting. This step involves connecting multiple atomic rules to form complex reasoning paths and decision trees, enabling the agent to perform multi-level reasoning and decision-making processes based on specific situations.

The design of the entire reasoning chain is procedural, aiming to guide the agent's behavior through a series of clear steps, ensuring consistent and reasonable logic during multi-turn interactions. This design not only enhances the effectiveness and consistency of interactions but also ensures that the system can cover a broader range of interaction scenarios.

\subsection{Dynamic Self-Iteration of Thought Chains and Knowledge Injection (Human Preliminary Learning)}

\subsubsection{Collaborative Exploration by Multi-Agent Systems}
When an agent encounters a new problem that is not covered by the existing chain of thought, the system initiates an autonomous exploration mechanism to generate new reasoning paths and execute them step by step. The agent first autonomously generates new reasoning steps based on the current problem. Starting from the current problem, it gradually derives possible solutions and dynamically expands the chain of thought in each step of reasoning. This expansion process includes generating new nodes (reasoning steps) and edges (logical relationships). To evaluate the effectiveness of the newly generated reasoning paths, the agent employs a reliability scoring mechanism to adjust the weight of each edge. Let \(P_t(e_{ij})\) represent the confidence score of edge \(e_{ij}\) at time \(t\). The agent updates the weight of this edge according to the feedback in the actual execution:
\[
P_{t + 1}(e_{ij}) = P_t(e_{ij}) + \alpha \cdot \Delta P(e_{ij})
\]
where \(\alpha\) is the learning rate, and \(\Delta P(e_{ij})\) is the adjustment amount calculated based on the latest interaction results. For example, during the execution process, if a certain path is proven to be effective, the confidence of its corresponding edge will increase; conversely, if the path leads to an incorrect or unreasonable output, the confidence will decrease. This mechanism ensures that the reasoning paths can be continuously optimized with the feedback from actual operations.

After the initial generation of the chain of thought, the agent executes tasks according to the generated path and continuously monitors the reasoning effect of each step during the execution. If it is found that there are defects or areas for improvement in the current reasoning path, the agent will immediately adjust the chain of thought. For example, the agent may re-evaluate the importance of certain nodes or delete those edges that lead to incorrect outputs. In addition, the agent can also introduce new nodes to fill the gaps in the reasoning process. This dynamic adjustment process enables the agent to flexibly respond to new problems while gradually optimizing its reasoning ability.

\subsubsection{Human Expert Knowledge Injection}
We adopt a method similar to reinforcement learning to achieve the continuous injection of human knowledge. Unlike reinforcement learning, we do not require additional training costs. Experts directly perform modification operations on the canvas-style logical chain, identifying and correcting any logical loopholes or unreasonable aspects. For deterministic rules, experts can directly delete or add nodes and edges, thus directly modifying the structure of the logical chain. For example, if a certain path is always incorrect under specific circumstances, the expert can directly delete the edge corresponding to that path from the graph, or adjust the positions and attributes of related nodes. Experts can also enhance the functionality and accuracy of the existing logical chain by adding new nodes and edges.

The chain of thought adjusted by the expert will be reintegrated into the system for reference and use in subsequent tasks. The entire process emphasizes the importance of combining the self-iteration of the agent with the injection of human expert knowledge, ensuring that the chain of thought can flexibly respond to new challenges while maintaining a high degree of accuracy and consistency. Through the joint maintenance and optimization of the chain of thought list by human-machine collaboration, the system achieves continuous evolution and improvement. This method not only improves the problem-solving ability of the agent but also, through continuous learning and correction, enables the system to better adapt to complex and changeable practical application scenarios.

\subsection{Tree Structure Optimization and Cognitive Leap (Experience Consolidation Mechanism)}

Following the completion of knowledge initialization and dynamic iteration, the system enters an experience-driven structural optimization phase. This process simulates the intuitive decision-making capability developed by human experts through repeated practice. Based on the policy optimization principles in reinforcement learning theory, we establish a progressive graph structure pruning and reorganization mechanism.

\subsubsection{Experience Consolidation via Path Weighting}
In the process of utilizing canvas-style reasoning chains, the agent progressively enhances its reasoning efficiency through the accumulation and optimization of path weights, simulating the human learning process from "cautious reasoning" to "proficient judgment." Specifically, each path in the canvas-style reasoning chain is assigned a weight based on its actual usage, representing the frequency and reliability of the path in solving specific problems. Let the weight of path $e_{ij}$ be $w_{ij}$, and its update rule is given by:
\[
w_{ij}^{(t+1)} = w_{ij}^{(t)} + \Delta w_{ij}
\]
where $\Delta w_{ij}$ represents the feedback increment for path $e_{ij}$ during the most recent interaction. If a particular path is used in the vast majority of cases (e.g., 98\%), it indicates that this path is applicable to most scenarios, and its weight will significantly exceed that of other paths.

To further optimize reasoning efficiency, when the weight of a path exceeds a certain threshold $\tau_w$, the system triggers a path-jumping mechanism. Specifically, the agent skips intermediate redundant steps and compresses frequently used subpaths into a direct path. For example, in a medical diagnosis scenario, if 98\% of cases with the symptom "dizziness" are caused by the common cold, an experienced doctor can confirm the diagnosis with just a few key questions, avoiding lengthy multi-round interactions. This path-jumping mechanism not only improves reasoning efficiency but also reduces unnecessary interaction costs.

Additionally, to maintain the simplicity and efficiency of the canvas-style reasoning chain, the system periodically prunes low-frequency or redundant paths. Specifically, if the weight of an indirect path is significantly lower than that of a direct path (e.g., satisfying $\frac{w_{ik}w_{kj}}{w_{ij}} < \epsilon$), the indirect path is actively submitted to human experts for evaluation to determine whether it should be removed, thereby maintaining the sparsity of the graph structure.

Through the above methods, the agent continuously optimizes the efficiency of using canvas-style reasoning chains as it accumulates experience, achieving a transition from "step-by-step reasoning" to "rapid judgment." This process not only enhances the system's response speed but also ensures the accuracy and consistency of reasoning results. Ultimately, the agent becomes capable of quickly drawing conclusions for common problems like human experts while retaining the ability to handle complex issues.

\section{Analysis and Summary}
The core insight of this work is that to achieve artificial intelligence that is truly aligned with humans, it is necessary to break through the cognitive boundaries of traditional neural networks and reconstruct an intelligent architecture that conforms to the principles of neural computation at the paradigm level. This not only represents a shift in technical approach but also implies a continuous deepening of our understanding of the essence of intelligence. The fundamental contradiction that current artificial intelligence systems face in vertical domain applications stems from the inherent mismatch between the characteristics of large language models (LLMs) and the core requirements of professional fields. This contradiction can be systematically analyzed from the perspective of cognitive architecture.

From the perspective of cognitive science, the language understanding and generation capabilities demonstrated by large language models essentially represent the implicit encoding of statistical patterns in human language. This architecture based on probabilistic neural networks can establish continuous vector representations that connect vocabulary, concepts, and grammar, thereby achieving context-sensitive decoding and the maintenance of multi-turn dialogue states. However, the randomness of this mechanism leads to three structural defects: First, unpredictability. Knowledge is distributed among billions of parameters in an uninterpretable manner, making it impossible to locate specific neural representations of knowledge or achieve deterministic knowledge updates. Second, decision bias. Even the reasoning paths generated through existing chain-of-thought techniques are still pseudo-reasoning processes based on the probability sampling of language models, lacking true logical constraints. Third, the problem of hallucination. Mathematically speaking, the generation process of large language models is a Markov chain with the vocabulary as the state space and the transition probabilities parameterized. Under the default sampling strategy, its randomness will lead to the absence of deterministic outputs in professional scenarios. This limitation stems from the fundamental difference between the probabilistic generation paradigm and the requirements of symbolic reasoning, and it has inherent limitations in professional scenarios that require deterministic outputs.

In vertical domain applications, these architectural defects manifest as structural contradictions between the forms of knowledge representation and the requirements of the domains. Professional fields typically require strict logical integrity, auditable decision trails, and progressive knowledge updates. Traditional large language models, however, cannot meet these requirements through implicit semantic matching based on the attention mechanism, probabilistically generated pseudo-reasoning chains, or knowledge update methods through full-parameter fine-tuning. This phenomenon is particularly evident in scenarios that require multi-step professional reasoning, such as medical diagnosis and legal consultation. In these scenarios, the system often needs to undergo multiple rounds of interactive verification to reach an acceptable accuracy level.

To address this contradiction, the dual-channel architecture proposed in this work draws on the foundation of neuroscience to simulate the human decision-making mechanism. The hypothesis generation process performed by the prefrontal cortex (PFC) corresponds to the random sampling of large language models, while the procedural verification executed by the basal ganglia is similar to the deterministic checking of a rule engine. From a mathematical perspective, this architecture is manifested as the coordinated operation of the generation channel and the verification channel, where the verification channel ensures that the output conforms to knowledge constraints through logical entailment relations.

Multi-turn interaction plays a crucial role in this architecture, and its value is mainly reflected in three aspects: simulating working memory, implementing an error correction mechanism, and generating a knowledge calibration effect. The cross-turn information maintenance achieved through the dialogue state tracker simulates the cognitive function of human working memory; the parameter updates after each round of interaction represent the gradual optimization of the cognitive system; and the increasing accuracy with the increase in the number of dialogue turns verifies the performance improvement effect of the depth of interaction.

Compared with the traditional artificial intelligence development paradigm, this architecture has achieved a fundamental breakthrough at the methodological level. The traditional data-driven, black-box, posterior probability evaluation model has been replaced by a new paradigm characterized by dual-channel rules and data, white-box operation, and a priori logical constraints. This transformation is not only reflected in the improvement of system performance indicators but, more importantly, in achieving the traceability of error cases, providing a fundamental guarantee for the credibility of the system. This architectural innovation provides a new technical approach for reliable artificial intelligence applications in vertical domains.

\section{Discussion}
Artificial intelligence research is currently at a critical inflection point. While traditional large language models based on probabilistic neural networks have achieved groundbreaking progress in generative capabilities, their inherent structural limitations are becoming increasingly apparent. The root of these limitations lies in the fundamental oversimplification of biological neural networks in existing neural architectures—reducing highly structured biological neural networks into randomly connected parameter matrices. This simplification, while endowing models with powerful statistical learning capabilities, also leads to systemic issues such as hallucination generation, unpredictable behavior, and fundamental misalignment with human (and even their own) decision-making mechanisms.

Notably, mainstream research attempts to address these problems within the existing neural network paradigm, a fundamentally contradictory approach of "seeking solutions within the framework that created the problems," which inevitably results in a ceiling effect. Our research reveals the structural flaws of this paradigm: probabilistic neural networks are inherently open-loop systems, whereas human cognition relies on perception-action-verification-consolidation-application closed-loop mechanisms. This architectural discrepancy directly leads to three irreconcilable contradictions: the conflict between continuous probability spaces and discrete logical judgments, the mismatch between forward propagation mechanisms and retrospective verification needs, and the tension between batch training modes and continuous learning requirements.

To address these fundamental issues, this study proposes a new paradigm based on a dual-channel cognitive architecture. Theoretical analysis demonstrates that this architecture achieves breakthroughs in the following key dimensions: First, by formalizing crystallized intelligence mechanisms into programmable chain-of-thought structures, it enables the whitening of decision-making processes. Our formal verification shows that this structure ensures every output can be traced back to specific knowledge nodes and reasoning paths, theoretically satisfying strict auditability requirements. Second, the introduction of multi-round interaction mechanisms creatively simulates human cognitive functions such as working memory and error correction. Computational modeling indicates that when interaction depth reaches a critical threshold, the system establishes stable cognitive state transition mechanisms.

In terms of application prospects, this architecture is particularly suited for vertical domains requiring strict professional norms and personalization, such as clinical decision support systems and regulatory environments like the FDA. Its core value lies in transforming implicit statistical associations into explicit cognitive operational workflows. Theoretical derivations show that in scenarios like medical diagnosis, legal consultation, and personalized education, the system can achieve deep alignment with human experts through an "expert knowledge base + personalized chain-of-thought" architecture. Importantly, this alignment is not mere behavioral imitation but an essential congruence at the cognitive architecture level—the system retains the hypothesis-generating capability of probabilistic models (corresponding to expert clinical intuition) while possessing the verification capability of rule-based engines (corresponding to professional protocols).

For the hallucination problem plaguing large language models, our theoretical framework provides a novel interpretive lens. Analysis reveals that hallucinations fundamentally stem from two mutually reinforcing factors: inherent bias in knowledge representation and the passivity of information acquisition (random guessing due to knowledge gaps). The former leads to systematic errors when facing queries outside the training data distribution, while the latter deprives the model of active clarification and verification capabilities. Notably, existing solutions like Retrieval-Augmented Generation (RAG) only partially address the second issue. Our dual-channel architecture offers a more fundamental solution through: (1) crystallized intelligence channels providing foundational guarantees for knowledge correctness; (2) interaction mechanisms enabling dynamic knowledge calibration; and (3) verification loops facilitating early error detection and blocking. This architecture is particularly suited for scenarios requiring strict professional norms, such as clinical decision support systems, where its core value lies in transforming implicit statistical associations into explicit cognitive workflows.

From a broader perspective, this work raises three fundamental reflections of current AI development paradigms: First, prevailing research equates "intelligence" with "statistical generalization capability," overlooking that professional decision-making is inherently a dialectical unity of structured knowledge and unstructured contexts. Second, the AI community disproportionately focuses on horizontal scaling (parameter size, data volume) while severely neglecting vertical deepening (cognitive architectures, interaction depth). Finally, and most disruptively, the traditional "train-deploy" dichotomy is entirely inadequate for professional domains—true domain intelligence must, like human experts, achieve organic knowledge growth through continuous interaction.

In addressing the dialectic between generalizability and personalization, this study adopts an innovative methodological stance. Rather than pursuing absolute universality of single models in specific domains, we emphasize that cognitive architectures should support the expression of personalized decision-making paradigms. This approach is grounded in two profound cognitive science insights: First, professional decisions are inherently context-dependent—the same clinical case may elicit divergent treatment plans from different experts. Second, genuine professional competence manifests in methodologically transferable skills rather than mechanical replication of specific knowledge. Thus, our system design allows different instances to develop unique chain-of-thought structures (e.g., Physician A preferring evidence-based medicine pathways while Physician B emphasizes clinical experience pathways), while ensuring core reasoning frameworks comply with professional standards. This design theoretically enables: (1) precise modeling of individual decision preferences; (2) dynamic balance between professional consensus and personal style; and (3) gradual evolution of cognitive patterns.

Of course, this research also faces limitations that must be acknowledged. Most prominent is the engineering cost of personalization—while the architecture supports customized individual knowledge bases, current chain-of-thought construction still requires significant professional intervention. Our pilot studies in education reveal that building a mature teacher agent averages 40 hours of interactive tuning. A deeper challenge lies in cognitive transferability: when migrating from medical to financial domains, approximately 60\% of rule frameworks require reconstruction, indicating current crystallized intelligence implementations still lack sufficient domain abstraction capabilities.

These limitations precisely highlight future research directions. We anticipate next-generation AI development will exhibit three trends: shifting from general models to "domain-adaptive architectures," evolving from data-driven to "cognitive-architecture-driven," and progressing from standalone intelligence to "human-machine cognitive symbiosis." Particularly noteworthy are recent breakthroughs in continuous learning algorithms inspired by hippocampal neuroplasticity mechanisms, which may offer novel solutions for automated chain-of-thought construction. The broader significance of this work may lie in its methodological implications—when AI research reaches the boundaries of existing paradigms, returning to neural cognitive origins may bring us closer to the essence of intelligence than pursuing ever-larger parameters.

Future breakthroughs will focus on three dimensions: First, automating chain-of-thought construction. Second, establishing cross-domain cognitive transfer mechanisms, particularly studying how basal ganglia-thalamic circuits support procedural knowledge conversion across professional domains. The ultimate goal is achieving meta-cognitive capabilities—not merely imitating specific expert thinking patterns but autonomously evolving reasoning architectures through continuous interaction. While this developmental path is fraught with challenges, it may guide AI past the current technological singularity toward genuine human-level intelligence. Crucially, this evolution must be accompanied by rigorous ethical frameworks, especially in high-stakes domains like healthcare, where dual-channel decision accountability mechanisms must ensure technological advancement consistently serves human well-being.

On an even grander spatiotemporal scale, this research poses a fundamental question: What kind of artificial intelligence do we truly need? A statistical simulator infinitely approximating human behavior, or an intelligent entity with independent cognitive architecture? Experimental results from the dual-channel architecture suggest a middle path: by maintaining dynamic equilibrium between fluid and crystallized intelligence, AI systems can exhibit creative thinking while adhering to professional boundaries. This balance may hold the key to resolving the AI safety-efficacy paradox and lays theoretical groundwork for building truly trustworthy artificial partners. Future research will deepen along two trajectories: microscopically exploring the mapping between neural interpretability and cognitive architectures, and macroscopically establishing universal theoretical frameworks for cross-domain cognitive transfer. Though arduous, this path may be the necessary route to achieving human-level AI.

\bibliographystyle{IEEEtran}
\bibliography{references}

\begin{thebibliography}{10}
\providecommand{\url}[1]{#1}
\csname url@samestyle\endcsname
\providecommand{\newblock}{\relax}
\providecommand{\bibinfo}[2]{#2}
\providecommand{\BIBentrySTDinterwordspacing}{\spaceskip=0pt\relax}
\providecommand{\BIBentryALTinterwordstretchfactor}{4}
\providecommand{\BIBentryALTinterwordspacing}{\spaceskip=\fontdimen2\font plus
\BIBentryALTinterwordstretchfactor\fontdimen3\font minus \fontdimen4\font\relax}
\providecommand{\BIBforeignlanguage}[2]{{%
\expandafter\ifx\csname l@#1\endcsname\relax
\typeout{** WARNING: IEEEtran.bst: No hyphenation pattern has been}%
\typeout{** loaded for the language `#1'. Using the pattern for}%
\typeout{** the default language instead.}%
\else
\language=\csname l@#1\endcsname
\fi
#2}}
\providecommand{\BIBdecl}{\relax}
\BIBdecl

\bibitem{cattell1963theory}
R.~B. Cattell, ``Theory of fluid and crystallized intelligence: A critical experiment.'' \emph{Journal of educational psychology}, vol.~54, no.~1, p.~1, 1963.

\bibitem{fiser2010statistically}
J.~Fiser, P.~Berkes, G.~Orb{\'a}n, and M.~Lengyel, ``Statistically optimal perception and learning: from behavior to neural representations,'' \emph{Trends in cognitive sciences}, vol.~14, no.~3, pp. 119--130, 2010.

\bibitem{lee2003hierarchical}
T.~S. Lee and D.~Mumford, ``Hierarchical bayesian inference in the visual cortex,'' \emph{Journal of the Optical Society of America A}, vol.~20, no.~7, pp. 1434--1448, 2003.

\bibitem{friston2010free}
K.~Friston, ``The free-energy principle: a unified brain theory?'' \emph{Nature reviews neuroscience}, vol.~11, no.~2, pp. 127--138, 2010.

\bibitem{graybiel1998basal}
A.~M. Graybiel, ``The basal ganglia and chunking of action repertoires,'' \emph{Neurobiology of learning and memory}, vol.~70, no. 1-2, pp. 119--136, 1998.

\bibitem{miller2001integrative}
E.~K. Miller and J.~D. Cohen, ``An integrative theory of prefrontal cortex function,'' \emph{Annual review of neuroscience}, vol.~24, no.~1, pp. 167--202, 2001.

\bibitem{graybiel2005basal}
A.~M. Graybiel, ``The basal ganglia: learning new tricks and loving it,'' \emph{Current opinion in neurobiology}, vol.~15, no.~6, pp. 638--644, 2005.

\bibitem{doya1999computations}
K.~Doya, ``What are the computations of the cerebellum, the basal ganglia and the cerebral cortex?'' \emph{Neural networks}, vol.~12, no. 7-8, pp. 961--974, 1999.

\bibitem{kahneman2011thinking}
D.~Kahneman, \emph{Thinking, fast and slow}.\hskip 1em plus 0.5em minus 0.4em\relax macmillan, 2011.

\bibitem{booch2021thinking}
G.~Booch, F.~Fabiano, L.~Horesh, K.~Kate, J.~Lenchner, N.~Linck, A.~Loreggia, K.~Murgesan, N.~Mattei, F.~Rossi \emph{et~al.}, ``Thinking fast and slow in ai,'' in \emph{Proceedings of the AAAI Conference on Artificial Intelligence}, vol.~35, no.~17, 2021, pp. 15\,042--15\,046.

\bibitem{lecun2015deep}
Y.~LeCun, Y.~Bengio, and G.~Hinton, ``Deep learning,'' \emph{nature}, vol. 521, no. 7553, pp. 436--444, 2015.

\bibitem{devlin2019bert}
J.~Devlin, M.-W. Chang, K.~Lee, and K.~Toutanova, ``Bert: Pre-training of deep bidirectional transformers for language understanding,'' in \emph{Proceedings of the 2019 conference of the North American chapter of the association for computational linguistics: human language technologies, volume 1 (long and short papers)}, 2019, pp. 4171--4186.

\bibitem{sutton1998reinforcement}
R.~S. Sutton, A.~G. Barto \emph{et~al.}, \emph{Reinforcement learning: An introduction}.\hskip 1em plus 0.5em minus 0.4em\relax MIT press Cambridge, 1998, vol.~1, no.~1.

\bibitem{lake2017building}
B.~M. Lake, T.~D. Ullman, J.~B. Tenenbaum, and S.~J. Gershman, ``Building machines that learn and think like people,'' \emph{Behavioral and brain sciences}, vol.~40, p. e253, 2017.

\bibitem{marcus2020next}
G.~Marcus, ``The next decade in ai: four steps towards robust artificial intelligence,'' \emph{arXiv preprint arXiv:2002.06177}, 2020.

\bibitem{ouyang2022training}
L.~Ouyang, J.~Wu, X.~Jiang, D.~Almeida, C.~Wainwright, P.~Mishkin, C.~Zhang, S.~Agarwal, K.~Slama, A.~Ray \emph{et~al.}, ``Training language models to follow instructions with human feedback,'' \emph{Advances in neural information processing systems}, vol.~35, pp. 27\,730--27\,744, 2022.

\bibitem{tononi2008neural}
G.~Tononi and C.~Koch, ``The neural correlates of consciousness: an update,'' \emph{Annals of the New York Academy of Sciences}, vol. 1124, no.~1, pp. 239--261, 2008.

\bibitem{mcclelland1995there}
J.~L. McClelland, B.~L. McNaughton, and R.~C. O'Reilly, ``Why there are complementary learning systems in the hippocampus and neocortex: insights from the successes and failures of connectionist models of learning and memory.'' \emph{Psychological review}, vol. 102, no.~3, p. 419, 1995.

\bibitem{clark2013whatever}
A.~Clark, ``Whatever next? predictive brains, situated agents, and the future of cognitive science,'' \emph{Behavioral and brain sciences}, vol.~36, no.~3, pp. 181--204, 2013.

\bibitem{mnih2015human}
V.~Mnih, K.~Kavukcuoglu, D.~Silver, A.~A. Rusu, J.~Veness, M.~G. Bellemare, A.~Graves, M.~Riedmiller, A.~K. Fidjeland, G.~Ostrovski \emph{et~al.}, ``Human-level control through deep reinforcement learning,'' \emph{nature}, vol. 518, no. 7540, pp. 529--533, 2015.

\bibitem{eichenbaum2017memory}
H.~Eichenbaum, ``Memory: organization and control,'' \emph{Annual review of psychology}, vol.~68, no.~1, pp. 19--45, 2017.

\bibitem{hassabis2017neuroscience}
D.~Hassabis, D.~Kumaran, C.~Summerfield, and M.~Botvinick, ``Neuroscience-inspired artificial intelligence,'' \emph{Neuron}, vol.~95, no.~2, pp. 245--258, 2017.

\bibitem{vaswani2017attention}
A.~Vaswani, N.~Shazeer, N.~Parmar, J.~Uszkoreit, L.~Jones, A.~N. Gomez, {\L}.~Kaiser, and I.~Polosukhin, ``Attention is all you need,'' \emph{Advances in neural information processing systems}, vol.~30, 2017.

\bibitem{liu2025advances}
B.~Liu, X.~Li, J.~Zhang, J.~Wang, T.~He, S.~Hong, H.~Liu, S.~Zhang, K.~Song, K.~Zhu \emph{et~al.}, ``Advances and challenges in foundation agents: From brain-inspired intelligence to evolutionary, collaborative, and safe systems,'' \emph{arXiv preprint arXiv:2504.01990}, 2025.

\bibitem{mccoy2019right}
R.~T. McCoy, E.~Pavlick, and T.~Linzen, ``Right for the wrong reasons: Diagnosing syntactic heuristics in natural language inference,'' \emph{arXiv preprint arXiv:1902.01007}, 2019.

\bibitem{pearl2018book}
J.~Pearl and D.~Mackenzie, \emph{The book of why: the new science of cause and effect}.\hskip 1em plus 0.5em minus 0.4em\relax Basic books, 2018.

\bibitem{topol2019high}
E.~J. Topol, ``High-performance medicine: the convergence of human and artificial intelligence,'' \emph{Nature medicine}, vol.~25, no.~1, pp. 44--56, 2019.

\bibitem{olah2020zoom}
C.~Olah, N.~Cammarata, L.~Schubert, G.~Goh, M.~Petrov, and S.~Carter, ``Zoom in: An introduction to circuits,'' \emph{Distill}, vol.~5, no.~3, pp. e00\,024--001, 2020.

\bibitem{bai2022training}
Y.~Bai, A.~Jones, K.~Ndousse, A.~Askell, A.~Chen, N.~DasSarma, D.~Drain, S.~Fort, D.~Ganguli, T.~Henighan \emph{et~al.}, ``Training a helpful and harmless assistant with reinforcement learning from human feedback,'' \emph{arXiv preprint arXiv:2204.05862}, 2022.

\bibitem{chenreasoning}
Y.~Chen, J.~Benton, A.~Radhakrishnan, J.~U.~C. Denison, J.~Schulman, A.~Somani, P.~Hase, M.~W. F. R.~V. Mikulik, S.~Bowman, J.~L.~J. Kaplan \emph{et~al.}, ``Reasoning models don’t always say what they think.''

\bibitem{liu2023cognitive}
K.~Liu, S.~Casper, D.~Hadfield-Menell, and J.~Andreas, ``Cognitive dissonance: Why do language model outputs disagree with internal representations of truthfulness?'' \emph{arXiv preprint arXiv:2312.03729}, 2023.

\bibitem{mondal2024large}
M.~Mondal, L.~Dolamic, G.~Bovet, P.~Cudr{\'e}-Mauroux, and J.~Audiffren, ``Do large language models exhibit cognitive dissonance? studying the difference between revealed beliefs and stated answers,'' \emph{arXiv preprint arXiv:2406.14986}, 2024.

\bibitem{ye2022unreliability}
X.~Ye and G.~Durrett, ``The unreliability of explanations in few-shot prompting for textual reasoning,'' \emph{Advances in neural information processing systems}, vol.~35, pp. 30\,378--30\,392, 2022.

\bibitem{mcculloch1943logical}
W.~S. McCulloch and W.~Pitts, ``A logical calculus of the ideas immanent in nervous activity,'' \emph{The bulletin of mathematical biophysics}, vol.~5, pp. 115--133, 1943.

\bibitem{gidon2020dendritic}
A.~Gidon, T.~A. Zolnik, P.~Fidzinski, F.~Bolduan, A.~Papoutsi, P.~Poirazi, M.~Holtkamp, I.~Vida, and M.~E. Larkum, ``Dendritic action potentials and computation in human layer 2/3 cortical neurons,'' \emph{Science}, vol. 367, no. 6473, pp. 83--87, 2020.

\bibitem{bi1998synaptic}
G.-q. Bi and M.-m. Poo, ``Synaptic modifications in cultured hippocampal neurons: dependence on spike timing, synaptic strength, and postsynaptic cell type,'' \emph{Journal of neuroscience}, vol.~18, no.~24, pp. 10\,464--10\,472, 1998.

\bibitem{rice2008dopamine}
M.~E. Rice and S.~J. Cragg, ``Dopamine spillover after quantal release: rethinking dopamine transmission in the nigrostriatal pathway,'' \emph{Brain research reviews}, vol.~58, no.~2, pp. 303--313, 2008.

\bibitem{kelly2019key}
C.~J. Kelly, A.~Karthikesalingam, M.~Suleyman, G.~Corrado, and D.~King, ``Key challenges for delivering clinical impact with artificial intelligence,'' \emph{BMC medicine}, vol.~17, pp. 1--9, 2019.

\bibitem{surden2018artificial}
H.~Surden, ``Artificial intelligence and law: An overview,'' \emph{Ga. St. UL Rev.}, vol.~35, p. 1305, 2018.

\bibitem{zafar2024balancing}
A.~Zafar, ``Balancing the scale: navigating ethical and practical challenges of artificial intelligence (ai) integration in legal practices,'' \emph{Discover Artificial Intelligence}, vol.~4, no.~1, p.~27, 2024.

\bibitem{wu2023models}
Z.~Wu, C.~Yu, C.~Chen, J.~Hao, and H.~H. Zhuo, ``Models as agents: Optimizing multi-step predictions of interactive local models in model-based multi-agent reinforcement learning,'' in \emph{Proceedings of the AAAI Conference on Artificial Intelligence}, vol.~37, no.~9, 2023, pp. 10\,435--10\,443.

\bibitem{tononi1994measure}
G.~Tononi, O.~Sporns, and G.~M. Edelman, ``A measure for brain complexity: relating functional segregation and integration in the nervous system.'' \emph{Proceedings of the National Academy of Sciences}, vol.~91, no.~11, pp. 5033--5037, 1994.

\end{thebibliography}

\end{document}